\documentclass[lettersize,journal]{IEEEtran}
\usepackage[switch]{lineno}
\usepackage{moreverb}

\usepackage{moreverb}
\usepackage[colorlinks,bookmarksopen,bookmarksnumbered,citecolor=red,urlcolor=red]{hyperref}

\usepackage{amssymb}
\usepackage{amsmath}
\usepackage{graphicx,subfigure}
\usepackage[ruled, linesnumbered]{algorithm2e}
\usepackage{nccbbb}
\usepackage{booktabs}
\usepackage{hyperref}
\usepackage{url}
\usepackage{multirow}
\usepackage{bigstrut}
\usepackage{hyperref}
\usepackage{bm}
\usepackage{color}
\usepackage[utf8]{inputenc}

\begin{document}

\title{See More, Match Better: Multi-Source Feature Fusion for Two-View Correspondence Learning}

\author{Xiaojie Li, Xin Jiang, Luanyuan Dai, Jinnan Yang, Yongdong Zhang,~\IEEEmembership{Fellow,~IEEE}, Zechao Li,~\IEEEmembership{Senior Member,~IEEE}
\thanks{Xiaojie Li, Luanyuan Dai, Jinnan Yang, and Zechao Li are with the School of Computer Science and Engineering, Nanjing University of Science and Technology, Nanjing 210094, China (e-mail: xiaojieli@njust.edu.cn; dailuanyuan@njust.edu.cn; yangjinnan@njust.edu.cn; zechao.li@njust.edu.cn).
\emph{(Corresponding Author: Zechao Li)}.}
\thanks{Xin Jiang is with the State Key Laboratory of Communication Content Cognition, People’s Daily Online, Beijing, 100733, China, and the School of Computer Science and Engineering, Nanjing University of Science and Technology, Nanjing 210094, China (e-mail:xinjiang@njust.edu.cn).}
\thanks{Yongdong Zhang is with the School of Information Science and Technology, University of Science and Technology of China, Hefei 230026, China (e-mail:~zhyd73@ustc.edu.cn).}
}

\maketitle
  
\begin{abstract}
Two-view correspondence learning aims to distinguish true correspondences~(inliers) from false ones~(outliers) in image pairs by leveraging their underlying differences. Existing methods mainly rely on coordinate-based geometric consistency.  \textcolor{red}{However}, they often struggle with pseudo-consistent outliers in scenes containing repetitive structures, textureless regions, or locally similar geometric patterns.
To address this limitation, we propose \textbf{TriMatch}, a two-stage multi-source feature fusion framework for two-view correspondence learning. Specifically, TriMatch jointly extracts geometric, texture semantic, and structural semantic features in Stage I. 
To bridge the gap between semantic and geometric features, the textural and structural semantic features are aligned with the geometric features through dedicated Texture-Geometric Alignment and Structural-Geometric Alignment modules, respectively.
We further introduce a Semantic-Guided Correspondence Modulation module, which modulates geometric features using semantic information to suppress geometrically plausible but semantically inconsistent correspondences. In Stage II, a Hierarchical Semantic-Enhanced Correspondence Refinement strategy progressively models correspondence dependencies and recalibrates multi-context feature responses, enabling more reliable inlier-outlier discrimination. Extensive experiments demonstrate the effectiveness, robustness, and generalization capability of TriMatch.
\end{abstract}


\section{Introduction}
\IEEEPARstart{E}{stablishing} reliable two-view correspondences is one of the fundamental problems in computer vision, playing a crucial role in tasks such as 3D reconstruction \cite{snavely2008modeling,schonberger2016structure}, visual localization \cite{sattler2018benchmarking, zhang2023epm}, and multi-view geometry \cite{wu2021computational, huang2024robust}.
A typical two-view correspondence learning pipeline consists of keypoint detection, descriptor extraction, and initial correspondence generation.
The feature extraction methods~\cite{lowe2004distinctive,detone2018superpoint} are commonly employed to detect and describe salient regions in images, followed by nearest-neighbor search in the descriptor space to establish an initial correspondence set. 
%
However, in real-world scenarios, factors such as repetitive structures, viewpoint changes, illumination variations, and textureless regions often introduce a large number of outliers into the initial correspondence set, which can significantly degrade the stability and robustness of downstream vision tasks. Therefore, accurately identifying and removing outliers remains a key challenge in feature matching.

\begin{figure}[t]
	\centering
	\includegraphics[width=1\columnwidth]{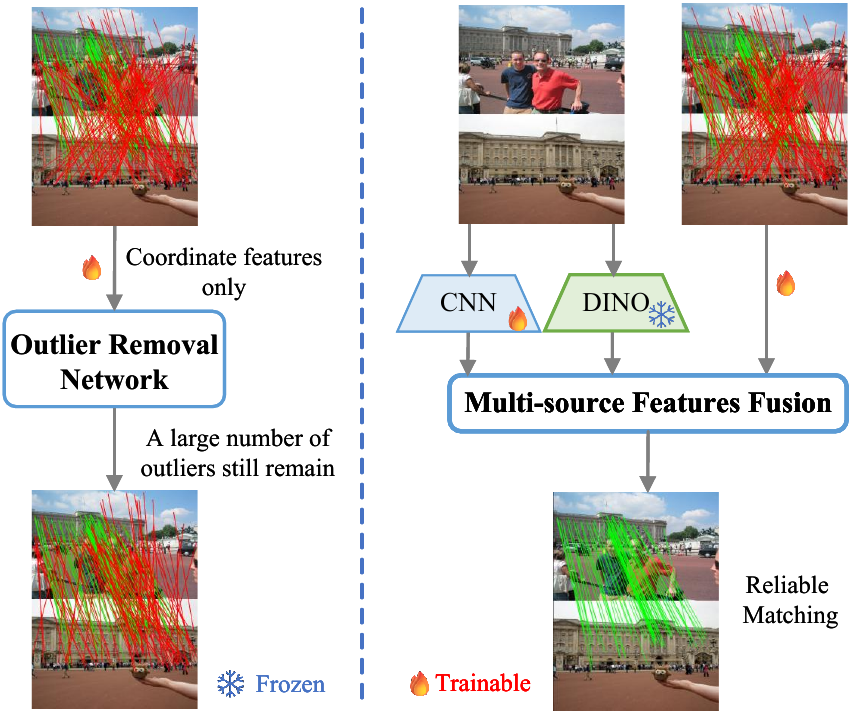}
     \vspace{-0.5cm}
	\caption{
Comparison between geometry-only modeling and multi-source feature fusion. Geometry-only methods (left) primarily rely on coordinate information and struggle to distinguish inliers from outliers in complex scenes, resulting in unreliable matching results. In contrast, our proposed TriMatch (right) integrates texture semantic features from CNNs, structural semantic features from DINOv2, and coordinate-based geometric features, which enhances the discriminability of correspondence representations, achieving more reliable outlier identification.
}
	\label{fig1}
\end{figure}

In recent years, learning-based methods \cite{yi2018learning, 2019Learning} have achieved remarkable progress by leveraging deep neural networks to model correspondence features, improving both outlier rejection and matching accuracy.
These methods typically treat geometric consistency as the primary constraint and further model neighborhood relationships among correspondences, either explicitly or implicitly, to enhance matching reliability. Representative approaches include graph neural networks (GNNs) \cite{sarlin2020superglue} and attention-based architectures, such as CLNet \cite{zhao2021progressive}, MS$^2$DGNet \cite{dai2022ms2dg}, NCMNet \cite{liu2024ncmnet}, MGNet \cite{luanyuan2024mgnet}, and BCLNet \cite{miao2024bclnet}. These methods construct neighborhood graphs over correspondences and exploit feature aggregation to capture consistency, thereby improving inlier identification.

However, most existing methods primarily rely on coordinate features to model geometric consistency. As a result, their discriminative capability is inherently limited by the expressiveness of coordinate-based geometric cues, particularly when randomly distributed outliers exhibit pseudo-consistent geometric patterns.
In challenging scenes with repetitive patterns, textureless regions, or similar local geometric layouts, such outliers may fall within the consistency neighborhoods formed by inliers. Consequently, they exhibit locally inlier-like geometric characteristics and become pseudo-consistent outliers. Under this interference, the reliability of neighborhood relationships becomes fragile, leading to degraded matching performance,  \textcolor{red}{as shown in the left part of Fig.~\ref{fig1}}.
This limitation raises a natural question: rather than relying solely on geometric features to model correspondence consistency, can semantic information inherent in the image be incorporated to learn more discriminative representations for accurate matching?

Fortunately, recent vision foundation models~\cite{he2016deep,oquab2023dinov2,dosovitskiy2020image} have demonstrated strong capabilities in visual correspondence and matching tasks. This opens up the possibility of leveraging image content as complementary evidence, moving beyond geometric reasoning based solely on coordinates.
Specifically, we decompose image content into two complementary semantic sources: texture semantics and structural semantics. Texture semantics provide local appearance-level cues around candidate correspondence regions, while structural semantics capture global layout-level context across broader image regions. For texture semantics, we employ a CNN~\cite{he2016deep} to extract fine-grained visual patterns. These cues help distinguish correspondences that share similar geometric configurations but differ in local appearance.
However, texture semantics may remain ambiguous in repetitive or textureless regions.
To alleviate this ambiguity, we further introduce structural semantics by leveraging DINOv2~\cite{oquab2023dinov2}, a self-supervised vision foundation model trained on large-scale data. It captures regional context, high-level semantic layouts, and cross-region structural relationships beyond coordinate-based geometric features. Such structural semantics help identify locally plausible but globally inconsistent correspondences, thereby improving discrimination in challenging scenes.

To this end, we propose \textbf{TriMatch}, a two-stage multi-source feature fusion framework that integrates geometric features, texture semantics, and structural semantics for robust two-view correspondence learning, as shown in the right part of Fig.~\ref{fig1}. Rather than simply stacking semantics features, TriMatch explicitly aligns heterogeneous semantic cues with geometric representations.
Specifically, in Stage I, coordinate-based geometric features are first extracted to provide an initial reliability prior for correspondence learning. Then, texture semantic features extracted by CNNs are aligned with geometric features through Texture-Geometric Alignment, while structural semantic features extracted by DINOv2 are aligned via Structural-Geometric Alignment. Subsequently, we introduce a Semantic-Guided Correspondence Modulation module to modulate geometric features using fused semantic information, thereby suppressing geometrically plausible but semantically inconsistent pseudo-consistent correspondences.
After enhancing individual correspondences with texture and structural semantics in Stage I, we propose a Hierarchical Semantic-Enhanced Correspondence Refinement strategy in Stage II to model higher-order dependencies among correspondences. This strategy progressively refines correspondence representations by reconstructing neighborhood relations, recalibrating relation-enhanced features, and propagating them through a U-shaped hierarchical architecture, resulting in more discriminative representations for reliable inlier–outlier separation.

In summary, the main contributions of this work are:
\begin{enumerate}
    \item We propose TriMatch, a two-stage multi-source verification framework that incorporates texture and structural semantics into geometric correspondence learning. 
    \item We propose a Texture-Geometric and Semantic-Geometric Alignment to align geometric and semantic features, along with Semantic-Guided Correspondence Modulation and Hierarchical Semantic-Enhanced Correspondence Refinement strategy to model dependencies among semantically aligned correspondences.
    \item Extensive experiments demonstrate that TriMatch achieves state-of-the-art performance, while maintaining favorable robustness and generalization across different datasets and descriptors.
\end{enumerate}

\section{Related Work}
\subsection{Traditional Outlier Removal Methods}
Classical outlier removal methods are mainly based on the sampling-and-consensus paradigm.
RANSAC~\cite{fischler1981random} is one of the earliest and most widely used methods, which estimates the optimal model by repeatedly sampling minimal subsets and evaluating model consistency. To improve sampling efficiency, PROSAC~\cite{chum2005matching} adopts a progressive sampling strategy based on correspondence confidence, which prioritizes high-quality matches and accelerates convergence. Its performance, however, depends on the quality of the initial correspondence ranking. USAC~\cite{raguram2012usac} further integrates multiple RANSAC-based strategies, including adaptive sampling, model verification, and local optimization, into a unified framework. MAGSAC~\cite{barath2020magsac++} formulates consensus evaluation from a probabilistic perspective, where the inlier threshold is marginalized to avoid manual tuning and achieve better stability under varying noise conditions.

In addition to sampling-based methods, researchers also utilize geometric constraints to approximate correspondence selection by modeling the consistency among correspondences.  
VFC~\cite{ma2014robust} represents correspondences as a smooth vector field and identifies inliers based on global motion consistency. GMS~\cite{bian2017gms} performs efficient correspondence filtering via local motion statistics within grid partitions, which is particularly effective under low inlier ratios.
For local structural modeling, LPM~\cite{ma2019locality} filters outliers by preserving neighborhood structural consistency, while LMR~\cite{ma2019lmr} further incorporates a regression mechanism to enhance discriminative capability on this basis.
COMR~\cite{xiao2021mining} formulates correspondence selection as a global consistency optimization problem, where correspondence confidence and geometric constraints are jointly modeled to improve robustness.
Despite these efforts, such methods still struggle in scenarios with high outlier ratios or heavily contaminated initial matches, where their performance degrades significantly.

\subsection{Learning-based Two-View Correspondence Learning}
In recent years, learning-based two-view correspondence methods have achieved significant progress, with the main objective of identifying inliers and removing outliers from an initial correspondence set. These methods can be broadly categorized into \textit{MLP-Based Correspondence Learning}, \textit{GNN-Based Correspondence Learning}, and \textit{Motion-Field-Based Correspondence Learning}. The MLP-Based Correspondence Learning methods introduce MLP as the backbone and treat two-view correspondence as a binary classification problem. LFGC~\cite{yi2018learning} employs MLPs with context normalization to process unordered correspondences to establish a foundational paradigm for correspondence learning. OANet~\cite{2019Learning} introduces order-aware filtering with DiffPool and DiffUnpool to aggregate both local and global contextual information, highlighting the importance of structural modeling. 

The GNN-Based Correspondence Learning methods construct graph structures over correspondences to capture geometric relationships. 
CLNet~\cite{zhao2021progressive} proposes a progressive consistency learning strategy, where dynamic graph construction enables hierarchical reasoning from local to global contexts and improves robustness under high outlier ratios. MS$^2$DGNet~\cite{dai2022ms2dg} performs progressive feature propagation over dynamically constructed sparse graphs. U-Match~\cite{li2023u} adopts hierarchical graph representation learning to compress redundant matches while preserving key structural information. NCMNet~\cite{liu2023progressive} models both intra-neighborhood context and inter-neighborhood interactions to exploit neighborhood consistency. DHM-Net~\cite{chen2024dhm} extends correspondences into a hypergraph space to capture group-wise relationships.

The Motion-Field-Based Correspondence Learning methods represent correspondences as motion vectors and learn inlier-outlier discrimination by modeling the smoothness of the resulting motion field. DeMo~\cite{lu2025deep} formulates global motion consensus within a learnable kernel and RKHS framework, enabling more stable and robust filtering. DeMatch++~\cite{zhang2025dematch++} models correspondences as a corrupted motion field and reduces the need for explicit regularization by decomposing the motion field, achieving efficient and implicit regularization.

Recently, VSFormer~\cite{liao2024vsformer} incorporates visual features into correspondence representations for pruning, offering a novel perspective on two-view correspondence; however, it mainly relies on simple feature-level fusion of visual information. In contrast, we explicitly decompose image content into two complementary semantic sources: texture semantics and structural semantics. Texture semantics capture fine-grained local appearance cues around correspondences, while structural semantics encode global layout and high-level contextual information using DINOv2~\cite{oquab2023dinov2}. Based on this decomposition, we design dedicated alignment modules to enable targeted interaction between semantic and geometric features, allowing our method to better suppress pseudo-consistent outliers that are locally plausible but globally inconsistent.

\subsection{Vision Foundation Models}

Since the introduction of ResNet~\cite{he2016deep}, vision foundation models have undergone rapid development. ViT~\cite{dosovitskiy2020image} successfully introduces the Transformer architecture into vision backbones and enables global feature modeling. ConvNeXt~\cite{liu2022convnet} revisits the design space of convolutional neural networks and achieves performance comparable to or exceeding that of ViT.
DINO~\cite{caron2021emerging} introduces a self-supervised learning framework that learns discriminative visual representations without manual annotations and demonstrates strong transferability across downstream tasks. Building on this line of research, DINOv2~\cite{oquab2023dinov2} further improves robustness and generalization by scaling up both training data and model capacity.
Benefiting from these advances, vision foundation models have shown strong potential in dense correspondence, visual localization, and segmentation tasks. 
Compared to geometric features, they provide richer structural and semantic cues, thereby enabling their application in correspondence modeling. In this paper, we introduce image semantic features and integrate them with geometric features to enhance matching robustness under challenging conditions.

%

\begin{figure*}[h]
    \centering
    \includegraphics[width=1\textwidth]{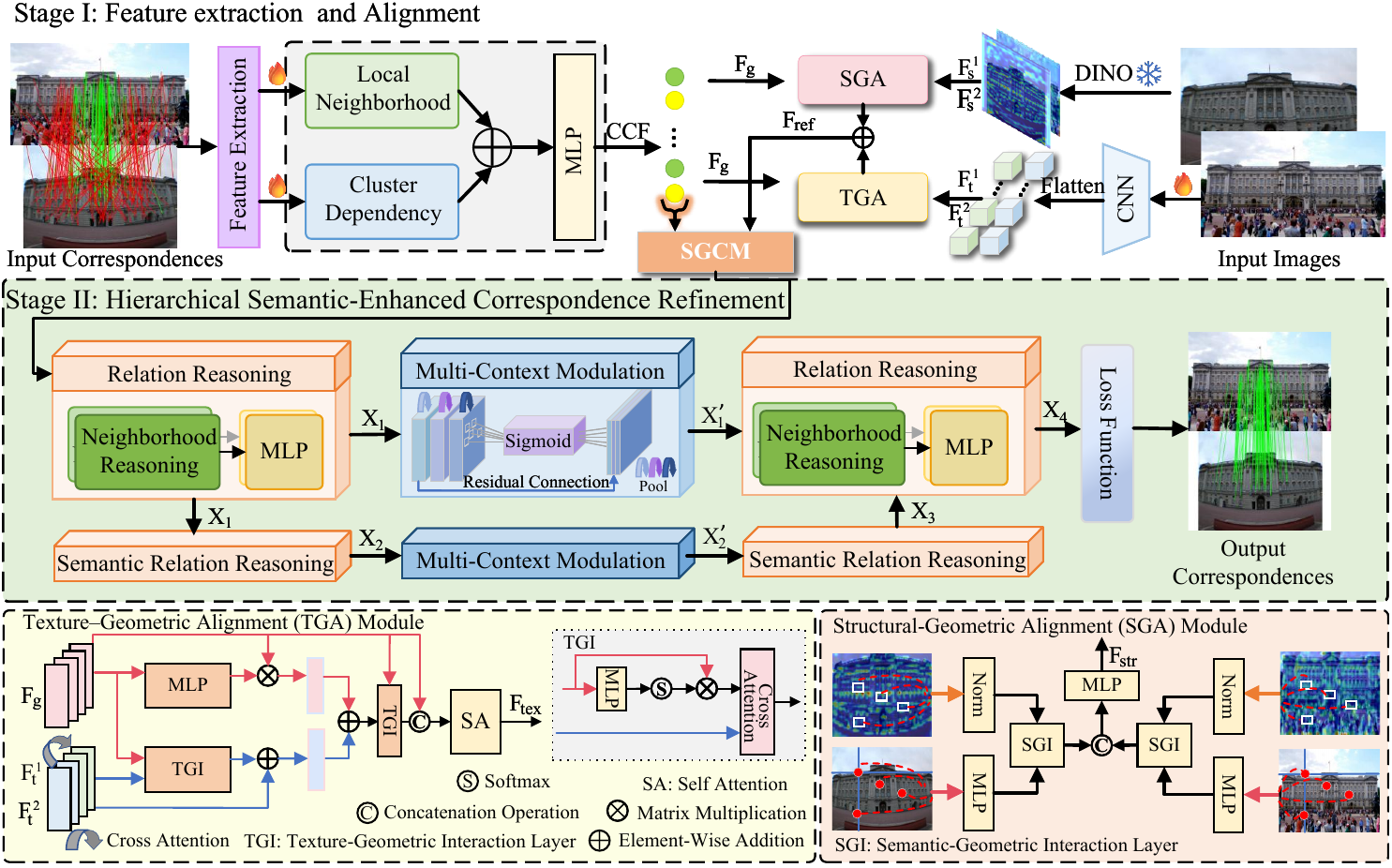}
    \caption{ The pipeline of the proposed TriMatch. The input is an initial correspondence set $C\in \mathbb{R}^{N \times 4}$. Subsequently, the network jointly models geometric, texture, and structural semantic features for correspondence learning, and finally outputs the probability set $P\in \mathbb{R}^{N \times 1}$ for inlier prediction. CCF is Coarse Correspondence Filtering. TGA and SGA denote the Texture-Geometric Alignment module and Structural-Geometric Alignment module, respectively. SGCM is a Semantic-Guided Correspondence Modulation module.
}
\label{fig2}
\end{figure*}

\section{Methodology}
TriMatch can be divided into two stages. The first stage focuses on extracting geometric, texture semantic, and structural semantic features and subsequently aligning and integrating them into a joint representation. In the second stage, we perform hierarchical semantic-enhanced correspondence refinement to further model dependencies among correspondences to obtain more discriminative correspondence representations.

\subsection{\textbf{Problem Formulation}}\label{PF}
Given a pair of images $\mathbf{I}$ and $\mathbf{I}'$, we first extract keypoints and descriptors using local feature detectors and descriptors~\cite{lowe2004distinctive, detone2018superpoint}, and establish an initial correspondence set by nearest-neighbor search.
The initial correspondence set is denoted as $\mathcal{C}=\{c_i\}_{i=1}^{N}\in\mathbb{R}^{N\times4}$, where each correspondence is defined as $c_i=(x_i,y_i,x_i',y_i')$.
Here, $(x_i,y_i)$ and $(x_i',y_i')$ denote the coordinates of the keypoints in two images $\mathbf{I}$ and $\mathbf{I}'$, respectively.

Based on the image pair and the initial correspondence set, TriMatch aims to predict the inlier probability for each candidate correspondence.
The overall network can be formulated as follows.
\begin{equation}
\mathbf{P}=\textrm{F}_{\mathrm{match}}(\mathbf{I},\mathbf{I}',\mathcal{C};\theta),
\end{equation}
where $\theta$ is the learnable parameter of TriMatch.
The output $\mathbf{P}=\{p_i\}_{i=1}^{N}$ is the predicted probability, $p_i \in [0,1]$ represents the probability that the correspondence $c_i$ is an inlier.

\subsection{\textbf{Stage I: Multi-Source Feature Alignment}}\label{efg}
Existing correspondence learning methods mainly rely on coordinate-based geometric consistency to model relationships among candidate correspondences. However, in challenging scenes with repetitive structures or similar local layouts, some randomly distributed outliers may fall into consistency neighborhoods and thus exhibit pseudo-consistency. This limitation makes geometry-only methods insufficient to reliably distinguish true inliers from pseudo-consistent outliers.

To improve correspondence discrimination, we introduce two complementary types of features: texture semantics and structural semantics. Texture semantics provide fine-grained local appearance cues to suppress coordinate-consistent but appearance-mismatched outliers, while structural semantics provide layout-aware contextual cues to identify locally plausible but globally inconsistent correspondences. Since these features differ in representation form and semantic granularity, directly concatenating or summing them may lead to semantic misalignment. Therefore, Stage I follows an extraction-alignment-modulation paradigm: we first extract multi-source features, then align texture and structural semantic features with geometric features through dedicated alignment modules, and finally use these semantic features to modulate geometric features, yielding a multi-source correspondence representation, as shown in Fig.~\ref{fig2}.



\subsubsection{\textbf{Geometric Feature Extraction}}
We extract coordinate-based geometric features as the basic reliability prior for correspondence learning. 
Given the initial correspondence set $\mathcal{C}=\{c_i\}_{i=1}^{N}\in\mathbb{R}^{N\times 4}$, we project it into a $D$-dimensional feature space using a $1\times1$ convolution, yielding $\mathbf{F}=[f_1,f_2,\ldots,f_N]\in\mathbb{R}^{N \times D}$.
To capture geometric relationships among correspondences, we model both local neighborhood consistency and cluster dependencies. For local modeling, we construct a K-Nearest Neighbor (K-NN) graph in the feature space. For the $i$-th correspondence, its neighborhood is denoted as $\{f_{ij}\}_{j=1}^{K}$. We define the edge feature as $e_{ij}=[f_i,\;f_i - f_j]$, where $[\cdot,\cdot]$ denotes concatenation.
The edge features are encoded by shared $1\times1$ convolutions and aggregated via max-pooling over the neighborhood to obtain the local geometric representation:
\begin{equation} \mathbf{F}_{\mathrm{local}}^{(i)}=\max_{j=1,\ldots,K} \mathrm{Conv}_{1\times1}(e_{ij}). 
\end{equation}

We employ a context aggregation operation~\cite{2019Learning} to capture long-range correspondence dependencies and enrich point-level geometric features with aggregated cluster information:
\begin{equation}
\mathbf{F}_{\mathrm{cluster}} = \mathcal{G}_{\mathrm{OA}}(\mathbf{F}),
\end{equation}
where $\mathcal{G}_{\mathrm{OA}}(\cdot)$ denotes the cluster context aggregation operation in OANet~\cite{2019Learning}.
Finally, we perform element-wise addition on the local and cluster features to obtain the unified feature representation $\mathbf{F}_\mathrm{uni}$:
\begin{equation}
\mathbf{F}_\mathrm{uni}=\text{MLP}\big(\mathbf{F}_{\mathrm{local}}+\mathbf{F}_{\mathrm{cluster}}\big),
\end{equation}
where $\mathrm{MLP}(\cdot)$ consists of a $1\times1$ convolution, Instance Normalization, Batch Normalization, and a ReLU activation.

\textbf{Coarse Correspondence Filtering.}
Based on $\mathbf{F}_\mathrm{uni}$, we predict a confidence score $\omega_i$ for each correspondence $c_i$.
All correspondences are ranked according to these scores, and the top-ranked $N_g=\lfloor \alpha N \rfloor$ correspondences are selected to form a refined correspondence subset $\mathcal{C'}$:
\begin{equation}
\mathcal{C'}=\mathrm{TopK}(C,\omega,N_g),
\end{equation}
where $\alpha$ is the sampling ratio, and $\mathrm{TopK}(\cdot)$ denotes the score-based filtering operator.
We then embed $\mathcal{C'}$ into a $D$-dimensional feature space using a $1\times1$ convolution, yielding the geometric feature $\mathbf{F}_\mathrm{g}\in\mathbb{R}^{N_g\times D}$. 
The correspondence in $\mathcal{C}'$ consists of a pair of matched keypoints from the image pairs. We further construct two image-specific geometric features, $\mathbf{F}_\mathrm{g}^1$ and $\mathbf{F}_\mathrm{g}^2$, from the coordinates associated with the first and second images, respectively.

\subsubsection{\textbf{Texture Semantics Extraction}}
Geometric features $\mathbf{F}_\mathrm{g}$ capture correspondence consistency; they may be insufficient when outliers are geometrically plausible but locally appearance-inconsistent with inliers.
To address this, we incorporate texture semantic features as complementary appearance cues, thereby improving correspondence discrimination.
Specifically, given a pair of images $\mathbf{I}, \mathbf{I}' \in \mathbb{R}^{3\times H \times W}$, we use a CNN-based encoder to extract texture semantic features, yielding $\mathbf{F}_\mathrm{t}^1=\mathrm{CNN}(\mathbf{I})$ and $\mathbf{F}_\mathrm{t}^2=\mathrm{CNN}(\mathbf{I}')$, where $\mathbf{F}_\mathrm{t}^1,\mathbf{F}_\mathrm{t}^2 \in \mathbb{R}^{D \times \frac{H}{4} \times \frac{W}{4}}$.
To enable cross-image interaction, we reshape the features into sequences: $\hat{\mathbf{F}}_\mathrm{t}^1=\mathrm{Flatten}(\mathbf{F}_\mathrm{t}^1)$ and $\hat{\mathbf{F}}_\mathrm{t}^2=\mathrm{Flatten}(\mathbf{F}_\mathrm{t}^2)$, where $\hat{\mathbf{F}}_\mathrm{t}^1,\hat{\mathbf{F}}_\mathrm{t}^2\in\mathbb{R}^{M\times D}$ and $M$ denotes the number of spatial tokens.

\textbf{Texture-Geometric Alignment.}
The geometric features are represented as sparse correspondence-level tokens, whereas texture semantic features are extracted as dense image-level representations on spatial grids. This difference in structural organization and semantic granularity makes direct joint modeling challenging. To bridge this gap, we propose a Texture-Geometric Alignment (TGA) module that aligns dense texture features with correspondence-level geometric features, enabling geometric tokens to adaptively aggregate relevant local texture information (Fig.~\ref{fig2}).

Given texture semantic features $\hat{\mathbf{F}}_\mathrm{t}^1$ and $\hat{\mathbf{F}}_\mathrm{t}^2$, we model cross-image joint representation via cross-attention:
\begin{equation}
\mathbf{F}_\mathrm{t}=\mathrm{Softmax}\left(\frac{\mathrm{MLP}(\hat{\mathbf{F}}_\mathrm{t}^1)\mathrm{MLP}(\hat{\mathbf{F}}_\mathrm{t}^2)^{\top}}{\sqrt{D}}\right)\mathrm{MLP}(\hat{\mathbf{F}}_\mathrm{t}^2),
\end{equation}
where $\mathbf{F}_\mathrm{t} \in \mathbb{R}^{M\times D}$.  
Considering that correspondence verification is performed at the correspondence level, it is necessary to associate dense texture tokens with geometric tokens. To this end, we design a Texture-Geometric Interaction (TGI) layer to establish geometry-to-texture interactions between $\mathbf{F}_\mathrm{g}\in\mathbb{R}^{N_g\times D}$ and $\mathbf{F}_\mathrm{t}\in\mathbb{R}^{M\times D}$. Specifically, we first compute a geometry-conditioned projection matrix 
$\mathbf{A}=\mathrm{Softmax}(\mathrm{MLP}(\mathbf{F}_\mathrm{g}))\in\mathbb{R}^{N_g\times M}$, where $\mathrm{MLP}(\cdot)$ maps each geometric token to an assignment distribution over the $M$ texture tokens. Using $\mathbf{A}$, the correspondence-level geometric features are projected into the texture-token space:
$\hat{\mathbf{F}}_\mathrm{g}=\mathbf{A}^{\top}\mathbf{F}_\mathrm{g}\in\mathbb{R}^{M\times D}$.

%

Based on the projected geometric feature, we apply cross-attention to establish geometry-to-texture interactions:
\begin{equation}
\mathbf{F}_{\mathrm{g \rightarrow  t}} = \mathrm{Softmax}\!\left(\frac{\mathrm{MLP}(\hat{\mathbf{F}}_\mathrm{g})\,\mathrm{MLP}(\mathbf{F}_\mathrm{t})^{\top}}{\sqrt{D}}\right)\mathrm{MLP}(\mathbf{F}_\mathrm{t}),
\end{equation}
where $\mathbf{F}_{\mathrm{g \rightarrow  t}}\in\mathbb{R}^{M \times D}$ denotes geometry-guided texture attention response.
Although $\mathbf{F}_{\mathrm{g \rightarrow t}}$ aggregates texture information under geometric guidance, the projection into the texture-token space may dilute the original geometric prior. To address this, we introduce a geometric compensation branch that explicitly preserves geometric cues. We first generate a projection matrix $\mathbf{P}_{\mathrm{g}\rightarrow \mathrm{t}}\in\mathbb{R}^{N_g \times M}$ from $\mathbf{F}{\mathrm{g}}$ using an MLP. The transpose of $\mathbf{P}_{\mathrm{g}\rightarrow \mathrm{t}}$ is then multiplied with $\mathbf{F}_{\mathrm{g}}$ to project geometric information into the texture-token space, yielding the compensation feature $\mathbf{F}^{p}_{\mathrm{g}\rightarrow \mathrm{t}}\in\mathbb{R}^{M \times D}$. We add this feature to $\mathbf{F}_{\mathrm{g}\rightarrow \mathrm{t}}$ to obtain the geometry-aware texture representation:
\begin{equation}
\mathbf{F}_{\mathrm{gt}}=
\mathbf{F}_{\mathrm{g \rightarrow t}} + \mathbf{F}^{p}_{\mathrm{g}\rightarrow \mathrm{t}},
\end{equation}
where $\mathbf{F}_{\mathrm{gt}}\in\mathbb{R}^{M\times D}$ captures both texture semantics and geometric constraints.

With $\mathbf{F}_{\mathrm{gt}}$, we perform texture-to-geometry interaction via TGI to project $\mathbf{F}_{\mathrm{gt}}$ back to the correspondence-token space for correspondence-level refinement.
\begin{equation}
\mathbf{B}=\mathrm{Softmax}(\mathrm{MLP}(\mathbf{F}_\mathrm{g\rightarrow t})),\quad 
\mathbf{F}_\mathrm{t\rightarrow g} = \mathrm{CA}\big(\mathbf{B}^{\top}\mathbf{F}_\mathrm{g\rightarrow t}, \mathbf{F}_\mathrm{g}\big),
\end{equation}
where $\mathrm{CA}(\cdot)$ denotes cross-attention and $\mathbf{B}\in\mathbb{R}^{M\times N_g}$. 
$\mathbf{F}_\mathrm{t\rightarrow g}\in\mathbb{R}^{N_g\times D}$ reassigns texture-enhanced information to correspondence-level tokens, allowing each candidate match to absorb relevant local appearance cues.

Finally, we concatenate $\mathbf{F}_\mathrm{t\rightarrow g}$ with the original geometric features $\mathbf{F}_\mathrm{g}$ to obtain the texture semantics $\mathbf{F}_{\mathrm{tex}}$:
\begin{equation}
\mathbf{F}_{\mathrm{tex}}=\mathrm{SA}\big([\mathbf{F}_\mathrm{t\rightarrow g}, \mathbf{F}_\mathrm{g}]\big),
\end{equation}
where $\mathrm{SA}(\cdot)$ denotes self-attention, $\mathbf{F}_{\mathrm{tex}}\in\mathbb{R}^{N_g\times D}$, and $[\cdot,\cdot]$ denotes concatenation.

\subsubsection{\textbf{Structural Semantics Extraction}}
Texture semantic features mainly focus on local appearance cues, which may still be ambiguous in repetitive or weakly textured regions. 
Therefore, broader structural semantic features are required to disambiguate visually similar regions and identify correspondences that are locally plausible but globally inconsistent with the scene structure.
To this end, we employ a pre-trained DINOv2~\cite{oquab2023dinov2} encoder as a frozen feature extractor to extract structural semantic features that encode global layout, regional context, and cross-region structural relationships.
Specifically, given images $\mathbf{I}$ and $\mathbf{I}'$, we extract  $\mathbf{F}_\mathrm{s}^1=\mathrm{DINO}(\mathbf{I})$ and $\mathbf{F}_\mathrm{s}^2=\mathrm{DINO}(\mathbf{I}')$, where $\mathbf{F}_\mathrm{s}^1,\mathbf{F}_\mathrm{s}^2\in\mathbb{R}^{H_s\times W_s \times D_s}$, $H_s$, $W_s$, and $D_s$ denote the spatial height, spatial width, and channel dimension, respectively.

\textbf{Structural-Geometric Alignment.}
To inject structural semantic features into correspondence representations, we design a Structural–Geometric Alignment (SGA) module (Fig.~\ref{fig2}). 
For clarity, we take the first image as an example. Let $\mathbf{F}_\mathrm{s}^1\in\mathbb{R}^{H_s\times W_s \times D_s}$ denote its structural semantic feature, and $\mathbf{F}_\mathrm{g}^1\in\mathbb{R}^{N_g\times D}$ denote the corresponding geometric feature.
These two features differ in both channel dimension and spatial organization; we first project $\mathbf{F}_\mathrm{s}^1$ into the same $D$-dimensional feature space as $\mathbf{F}_\mathrm{g}^1$ using a $1\times1$ convolution. We then apply bias-free normalization to alleviate distributional and scale discrepancies between heterogeneous features, thereby facilitating more stable structural-geometric alignment: $\hat{\mathbf{F}}_\mathrm{s}^1
=\gamma \odot \frac{\mathbf{F}_\mathrm{s}^1} {\sqrt{\mathrm{Var}(\mathbf{F}_\mathrm{s}^1)+\epsilon}}$,
where $\hat{\mathbf{F}}_\mathrm{s}^1\in\mathbb{R}^{H_s\times W_s \times D}$.
%

Next, we design a Structural-Geometric Interaction (SGI) layer to perform cross-modal interaction between structural semantic features and geometric features.
Specifically, to preserve the original structural layout as much as possible, we flatten 
$\hat{\mathbf{F}}_\mathrm{s}^1\in\mathbb{R}^{H_s\times W_s \times D}$ into a one-dimensional structural token sequence following a serpentine spatial order:
\begin{equation}
\mathbf{S}_\mathrm{s}^1[l,:]
=
\hat{\mathbf{F}}_\mathrm{s}^1[r,c_l,:],
\quad
r=\left\lfloor\frac{l}{W_s}\right\rfloor,
\end{equation}
where $l=0,\ldots,H_sW_s-1$, and
\begin{equation}
c_l=
\begin{cases}
l-rW_s, & r \ \mathrm{is\ even},\\
W_s-1-(l-rW_s), & r \ \mathrm{is\ odd}.
\end{cases}
\end{equation}
Here, $\mathbf{S}_\mathrm{s}^1\in\mathbb{R}^{H_sW_s\times D}$ denotes the reordered structural semantic sequence.

Subsequently, we apply one-dimensional linear interpolation along the token dimension of $\mathbf{S}_\mathrm{s}^1$, reducing its sequence length from $H_sW_s$ to $N_g$:
\begin{equation}
\mathbf{Q}_\mathrm{s}^1
=
\mathcal{I}_{N_g}
\left(
\mathbf{S}_\mathrm{s}^1
\right)
\in \mathbb{R}^{N_g\times D},
\end{equation}
where $\mathcal{I}_{N_g}(\cdot)$ denotes linear interpolation that adjusts the number of structural tokens to $N_g$. 
$\mathbf{Q}_\mathrm{s}^1$ serves as the query, while the geometric feature $\mathbf{F}_\mathrm{g}^1$ is projected into key and value embeddings:
$\mathbf{K}_\mathrm{g}^1,\mathbf{V}_\mathrm{g}^1
=
\mathrm{Conv}_{kv}(\mathbf{F}_\mathrm{g}^1)$,
where $\mathbf{K}_\mathrm{g}^1,\mathbf{V}_\mathrm{g}^1\in\mathbb{R}^{N_g\times D}$.
We then reshape $\mathbf{Q}_\mathrm{s}^1$, $\mathbf{K}_\mathrm{g}^1$, and $\mathbf{V}_\mathrm{g}^1$ into multi-head representations, \emph{i.e.,} $\mathbf{Q}_\mathrm{s}^1, \mathbf{K}_\mathrm{g}^1,\mathbf{V}_\mathrm{g}^1\in\mathbb{R}^{N_g \times h\times d}$ with $D=h\cdot d$. 
To improve numerical stability, the query and key are normalized in a head-wise manner:
$
\mathbf{Q}_{\mathrm{s},i}^1=\frac{\mathbf{Q}_{\mathrm{s},i}^1}{\|\mathbf{Q}_{\mathrm{s},i}^1\|}$ and
$\mathbf{K}_{\mathrm{g},i}^1=\frac{\mathbf{K}_{\mathrm{g},i}^1}{\|\mathbf{K}_{\mathrm{g},i}^1\|}, i=1,\dots,h.$
For each attention head, the structural semantic query attends to the geometric keys and aggregates the corresponding value features:
\begin{equation}
\mathbf{O}_\mathrm{s}^1 = 
\mathrm{Concat}_{i=1}^{h} 
\left( 
\mathrm{Softmax}
\left( 
\mathbf{Q}_{\mathrm{s},i}^1 
(\mathbf{K}_{\mathrm{g},i}^1)^\top 
\cdot \tau 
\right) 
\mathbf{V}_{\mathrm{g},i}^1 
\right),
\end{equation}
where $\tau$ is a learnable scaling parameter. 
The same operation is applied to the second pair of structural semantic features $\mathbf{F}_\mathrm{s}^2$ and geometric features $\mathbf{F}_\mathrm{g}^2$ to obtain $\mathbf{O}_\mathrm{s}^2$. Finally, we concatenate $\mathbf{O}_\mathrm{s}^1$ and $\mathbf{O}_\mathrm{s}^2$, and feed the result into an MLP to obtain the structural semantic features $\mathbf{F}_{\mathrm{str}}\in\mathbb{R}^{N_g \times D}$.

\subsubsection{\textbf{Semantic-Guided Correspondence Modulation}}
The $\mathbf{F}_{\mathrm{tex}}$ and $\mathbf{F}_{\mathrm{str}}$ encode complementary information from texture and structural perspectives, respectively. To suppress pseudo-consistent outliers with semantic information, we design a Semantic-Guided Correspondence Modulation module, whose core idea is to generate modulation weights from semantic features and apply them to the geometric features, enabling semantic features to guide geometric feature refinement.

Specifically, we first generate a geometry-aware weighting map $Z$ from $\mathbf{F}_\mathrm{g}$ to emphasize more reliable components in the geometric features:
\begin{equation}
\mathbf{Z} = \mathrm{Softmax}\left(\mathrm{MLP}(\mathbf{F}_\mathrm{g}) \odot \mathrm{MLP}(\mathbf{F}_\mathrm{g})\right),
\end{equation}
where $\odot$ denotes element-wise multiplication.
Then, we fuse $\mathbf{F}_{\mathrm{tex}}$ and $\mathbf{F}_{\mathrm{str}}$ via element-wise addition to obtain a unified semantic feature $\mathbf{F}_{\mathrm{sem}}$.
To modulate the geometric features with semantic information, $\mathbf{F}_{\mathrm{sem}}$ is fed into an MLP followed by a sigmoid function to generate semantic weights, which are further transformed into a residual form.
Based on the geometry-aware weighting map and the semantic weights, we refine the geometric features as:
\begin{equation}
\mathbf{F}_{\mathrm{ref}}
=
\mathbf{F}_{\mathrm{g}}
+
\left(1+\sigma\left(\mathrm{MLP}(\mathbf{F}_{\mathrm{sem}})\right)\right)
\odot
\left(\mathbf{Z}\odot\rho(\mathbf{F}_{\mathrm{g}})\right),
\end{equation}
where $\sigma(\cdot)$ denotes the sigmoid function, and $\rho(\cdot)$ is implemented by a $1\times1$ convolution followed by normalization and nonlinear activation. $\mathbf{Z}$ highlights reliable components in the transformed geometric features $\rho(\mathbf{F}_{\mathrm{g}})$, while $1+\sigma(\mathrm{MLP}
(\mathbf{F}_{\mathrm{sem}}))$ provides semantic modulation weights.

\subsection{\textbf{Stage II: Hierarchical Semantic-Enhanced Correspondence Refinement}}

In Stage I, we obtain the multi-source feature $\mathbf{F}_{\mathrm{ref}}$ to improve the discriminative ability of correspondences. However, while $\mathbf{F}_{\mathrm{ref}}$ enriches individual correspondences with semantic information, the dependencies among correspondences are not sufficiently modeled. Therefore, we propose a Hierarchical Semantic-Enhanced Correspondence Refinement strategy in Stage II to refine the correspondences by jointly modeling correspondence-level features and inter-correspondence semantic-geometric consistency, resulting in more discriminative correspondence representations, as illustrated in Fig.~\ref{fig2}.

Specifically, we introduce a hierarchical refinement architecture with U-shaped connections, which preserves structural cues from shallow layers while enabling deeper layers to capture broader contextual dependencies.
Within this architecture, a Relation Reasoning (RR) module reconstructs neighborhood relations based on $\mathbf{F}_{\mathrm{ref}}$, allowing dependencies among semantically enhanced correspondences to be explicitly modeled. In addition, a Multi-Context Modulation (MCM) module adaptively recalibrates the relation-enhanced features by integrating local details, global context, and salient discriminative responses.
Through this design, the model progressively refines correspondence representations by combining dependency modeling, multi-context modulation, and hierarchical feature propagation, producing more discriminative features for subsequent inlier prediction. 
The overall process can be formulated as:
\begin{equation} 
\begin{aligned} 
& \mathbf{X}_1 = \phi_1(\mathbf{F}_{\mathrm{ref}}), \quad \mathbf{X}'_1 = \varphi_1(\mathbf{X}_1), \\ 
& \mathbf{X}_2 = \phi_2(\mathbf{X}_1), \quad \mathbf{X}'_2 = \varphi_2(\mathbf{X}_2), \\ 
& \mathbf{X}_3 = \phi_3(\mathbf{X}'_2), \quad \mathbf{X}_4 = \phi_4(\mathbf{X}_3) + \mathbf{X}'_2,
\end{aligned} 
\end{equation}
where $\phi_l(\cdot)$ denotes the $l$-th RR module, which consists of a KNN-based relation layer \cite{zhao2021progressive} and an MLP to reconstruct neighborhood relations and model dependencies among correspondences.
$\varphi_l(\cdot)$ denotes the MCM module, which integrates multi-context cues to recalibrate correspondence features.

For the details of the MCM module, we take the first layer as an example.
The feature is modeled from three complementary views, including local, average-pooled, and max-pooled, to extract diverse contextual cues:
\begin{equation}
\mathbf{X}_{\mathrm{loc}}=\phi_{\mathrm{loc}}(\mathbf{X}_1),\quad
\mathbf{X}_{\mathrm{avg}}=\phi_{\mathrm{avg}}(\mathbf{X}_1),\quad
\mathbf{X}_{\mathrm{max}}=\phi_{\mathrm{max}}(\mathbf{X}_1).
\end{equation}
These views are then adaptively aggregated to recalibrate the original feature:
\begin{equation}
\mathbf{X}'_1 = \mathrm{MLP}\big(\sigma\big(\mathbf{X}_{\mathrm{loc}} + \mathbf{X}_{\mathrm{avg}} + \mathbf{X}_{\mathrm{max}}\big)\big) + \mathbf{X}_1,
\end{equation}
where $\sigma(\cdot)$ is the sigmoid function.
%
The final output, $\mathbf{X}_4$, serves as a refined and discriminative correspondence representation for subsequent inlier prediction.

\subsection{\textbf{Loss Function}}\label{LF}
To optimize TriMatch, we employ a hybrid loss function comprising a matrix regression loss \( \mathcal L_e \) and a classification cross-entropy loss \( \mathcal L_c \). The matrix regression loss \( \mathcal L_e \) is designed to evaluate the geometric discrepancy between the ground-truth essential matrix \( {E}^{\ast} \) and the model's predicted estimate \(\hat{{E}}\).
{\setlength{\abovedisplayskip}{4pt}
 \setlength{\belowdisplayskip}{4pt}
\begin{equation}\label{loss}
\mathcal L = \mathcal L_c + \beta \mathcal L_e({E}^{\ast}, \hat{{E}}),
\end{equation}}
where $\beta$ is a hyper-parameter to balance $\mathcal L_c$ and $\mathcal L_e$.

\section{Conclusion}
In this paper, we propose TriMatch, a two-stage multi-source feature fusion framework for two-view correspondence learning. In stage I, TriMatch integrates geometric, texture, and structural semantic features through dedicated alignment and semantic-guided modulation modules, enabling more discriminative correspondence representations. In Stage II, a Hierarchical Semantic-Enhanced Correspondence Refinement strategy is further introduced to progressively model correspondence dependencies and improve inlier-outlier discrimination. 
Extensive experiments on multiple benchmarks demonstrate that TriMatch achieves state-of-the-art performance, with strong robustness and generalization under challenging scenarios.

\bibliographystyle{IEEEtran}
\bibliography{ref}

\end{document}